\documentclass[11pt]{article}
\usepackage{eacl2017}
\usepackage{times}
\usepackage{url}
\usepackage{latexsym}

\eaclfinalcopy % Uncomment this line for the final submission
\def\eaclpaperid{...} %  Enter the acl Paper ID here

\title{Rule-Based Spanish Morphological Analyzer \\Built From Spell Checking Lexicon}

\author{Natalie Ahn \\
  University of California, Berkeley \\
  {\tt natalieahn@berkeley.edu} \\}

\date{}

\begin{document}
\maketitle
\begin{abstract}
Preprocessing tools for automated text analysis have become more widely available in major languages, but non-English tools are often still limited in their functionality. When working with Spanish-language text, researchers can easily find tools for tokenization and stemming, but may not have the means to extract more complex word features like verb tense or mood. Yet Spanish is a morphologically rich language in which such features are often identifiable from word form. Conjugation rules are consistent, but many special verbs and nouns take on different rules. While building a complete dictionary of known words and their morphological rules would be labor intensive, resources to do so already exist, in spell checkers designed to generate valid forms of known words. This paper introduces a set of tools for Spanish-language morphological analysis, built using the COES spell checking tools, to label person, mood, tense, gender and number, derive a word's root noun or verb infinitive, and convert verbs to their nominal form.
\end{abstract}

\section{Introduction}

Methods and tools for extracting information from text have come a long way. In at least a handful of major languages, there are now widely-used pre-trained tools available for part speech tagging, parsing, and named entity recognition, such as the Stanford CoreNLP toolkit,~\cite{Manning:14}. These tools work well for bag-of-words approaches to analyzing documents' topics and tones, and for limited forms of relation extraction. However, if researchers seek to use more complex linguistic structure to extract more detailed information from text, they may need to identify word features beyond their frequency or simple grammatical roles. Researchers may need to extract information about the state of certain actions, from verbs' person, mood, and tense, or label noun gender and number to aid coreference resolution.

Efforts to improve these latter tasks are ongoing, and tools to implement them tend to be more developed for English than for other languages. Researchers seeking to fill in additional steps on their own in other languages often face challenges even when it comes to basic text processing steps. Annotated corpora tend to be fewer and smaller in languages other than English, which can lead to even greater data sparsity in morphologically rich languages~\cite{LeRoux:12,Labaka:07}. Multilingual toolkits also often have more limited resources for non-English languages. For instance, Python's Natural Language Toolkit (NLTK) contains a stemmer but no lemmatizer for Spanish words, yet its interface to the Spanish component of Open Multilingual WordNet~\cite{Gonzalez-Agirre:12} only recognizes valid infinitives for verbs, but not conjugated verbs or truncated stems.

Yet some foreign languages are especially well suited to extracting or deconflicting additional pieces of information about word features and forms. For instance, Spanish language verb conjugations follow a much richer and more consistent set of morphological rules than English, as do gender and number in Spanish noun forms. A rule-based approach to word form analysis can be easier to develop with limited resources than a machine learning-based approach. General knowledge of a language's morphological patterns combined with general lexical resources may be sufficient to label detailed word features useful to a variety of applications.

Yet we still need some lexical knowledge mapped to morphological patterns; understanding of common rules is not enough. Spanish verbs have clear conjugation rules, but there are many irregular verbs that each have their own set of rule-based forms. Nominal forms of verbs have recognizable endings, but there are multiple common transformations and each verb may use only one of those options. Gender and number follow very simple rules, but there are still plenty of exceptions, e.g. male nouns that end in \emph{-a} and female nouns that end in \emph{-o}. A dictionary of known words and their permitted transformations would be labor intensive to build, but similarly structured resources already exist. Specifically, spell checking software often combines dictionaries of known words with a set of morphological rules permitted for each root. While the purpose of spell checking is simply to determine a token's validity in a given language, the existing structure can be augmented with features assigned to specified morphological rules to facilitate additional word form analysis.

This project builds on the COES Spanish Language Tools for spell checking~\cite{Rodriguez:96}. A single researcher was able to efficiently transform that resource into a set of lexicon, morphological rules, and Python tools for identifying verb person, mood, and tense, noun or adjective gender and number, as well as for lemmatization and related word transformations. This paper serves not only as an introduction to a resource that may be valuable to others working with Spanish text, but also as an example of what might be done in other languages with limited project-specific resources, where general language resources of this type are available.

This paper proceeds as follows. The first two sections describe related work and the original COES resource on which this project was built. The paper then describes what was done to transform and augment the COES materials for use in word form analysis, and defines the resulting components of the project's materials and tools. The paper then presents a set of evaluations using test data from the CoNLL 2009 shared task, which contained a Spanish language annotated corpus for event extraction derived from the AnCora project. The new tools label gender, number, person, mood and tense, and lemmatize verbs to their infinitive form, with over 90\% accuracy on the test set.

\section{Related Work}

As mentioned above, the Stanford CoreNLP toolkit now provides Spanish language models for its most popular tools, including its part of speech tagger and parser, which tag verbs with mood and tense information based on the AnCora project's tagset. The tags do not distinguish speaker person or number, nor does CoreNLP provide Spanish coreference resolution or gender and number features for nouns or adjectives.

There are several annotated corpora for Spanish dependency parsing that include morphological word features. The Universal Dependencies project~\cite{McDonald:13} provides corpora in many languages, and the Spanish version is annotated with gender, number, person, mood, and tense information. The corpus could be used to train a machine learning-based feature extractor, as long as the training corpus covers the same terms as the researcher's domain. Since no rules are specified about the relationships between word forms, however, conjugated words cannot be traced back to their infinitive lemma, without additional knowledge.

Another resource that offers morphological analysis for Spanish text is FreeLing, an open source C++ library of language analysis tools covering Spanish and several other European languages~\cite{Atserias:06}. It focuses on detecting specialized expressions and quantities, including proper nouns, multiword expressions, word number, date/time information, percentages and ratios. FreeLing's \emph{morfo} tool applies a cascade of specialized processors to detect each of these features using dictionary lookup and suffix handling, along with quantitative expression recognizers. The Spanish and Catalan morphological dictionaries are hand built and are smaller than those FreeLing contains for other languages. The Spanish version contains the 6,500 most frequent open-category lemmas in the language, which correspond to 81,000 forms and are expected to cover at least 80\% of open-category tokens on unrestricted text. In contrast, the COES dictionary described in the next section contains over 50,000 root words and over ten times that many forms.

In an effort most similar to our own, the \emph{hunmorph} project extended the functionality of the spellchecker MySpell to use its resources for stemming and morphological analysis, applied to Hungarian and several related languages~\cite{Tron:05}. The authors argued that compiling dictionaries and coding the morphology of a language are such labor-intensive tasks, that there is great benefit from reusing language-specific resources assembled for spell checking across other tasks involving word-level analysis. Their work leverages the structure of MySpell and Ispell lexical resources, which assign morphological rules to specific roots in the lexicon, in order to avoid spurious hits from guessing roots based on general suffix patterns~\cite{Nemeth:04}. The same features are present in the COES resources discussed below, which is integrated into the Ispell tool for Spanish language text.

\subsection{COES Spell Checking Resource}

This project builds on the COES Spanish Language Tools (COES Herramientas para Procesamiento de Lenguaje Natural en Español), developed at the Universidad Politécnica de Madrid (UPM) and the Universidad Carlos III de Madrid~\cite{Rodriguez:96}. COES contains a dictionary and set of grammatical rules, designed for assessing the correctness of documents written in Spanish, which have been integrated into the International Ispell spell-checking program. The resource is available under a GNU General Public License. The version used in this project is v. 1.11, dated November 2010, the latest GPL version available as of January 2017.

The tools come with a makefile to generate a complete hash file of all derived word forms, which requires about 50Mb of paging space and 100Mb of temporary disk space to build, according to the COES website. The dictionary itself is only 4Mb and the rule file much smaller. While the resulting hash file contains only word forms with no feature information, the grammatical rule file does contain notes with partial information about the verb and noun forms represented by certain groups of rules.

For instance, the flag ``V'' in the COES rules file denotes the set of conjugation rules for regular verbs. Root words (mainly infinitives, but not always) in the dictionary files are marked with a ``/V'' if the regular verb conjugation rules should be applied to them. The ``V'' section of the rules file contains subheads with notes that indicate the mood and/or tense of each section of rules. Each rule is written on a separate line, containing a regular expression to match to the ending of a root to which the rule may be applied, the portion of the ending to be removed and the replacement ending to create the given form.

\begin{figure}[h]
\begin{center}
\begin{tabular}{|l|}
\hline
  \\
flag *V:     \# Verbos de todas las conjugaciones \\
    regulares ''amar'' PRESENTE \\
    A R		\textgreater -AR, O		\# amar amo \\
    \lbrack \^{} CG \rbrack E R \textgreater -ER, O		\# comer como \\
    C E R	\textgreater -CER, ZO		\# vencer venzo \\
    G E R	\textgreater -GER, JO		\# coger cojo \\
    . . . \\
 \\
flag *S:	\# Plural \\
    \lbrack AEIOU'A'E'O \rbrack \textgreater	S		\# vaca vacas \\
    \lbrack 'U'IDJLMRY \rbrack \textgreater	ES		\# tab'u tab'ues \\
    . . . \\
 \\
\hline
\end{tabular}
\end{center}
\caption{\label{font-table} Sample contents of COES \emph{espa\~{}nol.aff} rule file. }
\end{figure}

Figure 1 shows a sample of the original COES rule file. The first section shows regular verb forms for the present tense (``\emph{PRESENTE}'' noted in the section comment), indicative mood (not noted), for the first person (not noted), for infinitives ending in \emph{-ar}, then \emph{-er} (depending on the preceding consonant). The second segment shown in Figure 1 contains forms for plural nouns, with the number (``\emph{Plural}'') but not gender of the form indicated in the comment above the rules. Accented characters are designated with a single quote before the relevant letter.

\section{Development of Morphological Analyzer}

To construct this project's rule set, we first automatically parsed the COES rule file into a spreadsheet, separating root and morphological endings, extracting any mood or tense information from the comments above rule groups (e.g. searching for words like ``\emph{presente}'' or ``\emph{futuro}''). Morphological categories were then filled in by hand in the new rules file as needed. Speaker was not indicated in the COES rule file, but morphological rules were generally listed in consistent speaker order (e.g. first, then second, then third person, singular then plural) so that these features could be quickly inferred. In general, the ability to assign features to already structured morphological rules, as opposed to assigning them to each individual root word in the lexicon, made the process of adding information much easier and more straight-forward than starting from scratch.

Noun gender had to be assigned from scratch, initially by populating the new rules spreadsheet with this feature based on morphological endings (e.g. male for \emph{-o}, female for \emph{-a}). Some rules needed to be split so that more specific endings could be assigned different features (e.g. \emph{-d} and \emph{-z} endings are female, other consonants male). Then additional rules were created for nouns that violate standard morphological rules, such as male words ending in \emph{-a} and female words ending in \emph{-o}. We referred to Spanish language learning resources online for common lists of these irregular terms\footnote{About.com, ``Words That Break the Gender `Rule'.'' \\ http://spanish.about.com/cs/grammar/a/genderreversal.htm, accessed January 15, 2017. \\ SpanishDict, ``Masculine and Feminine Nouns.'' \\ http://www.spanishdict.com/guide/masculine-and-feminine-nouns, accessed January 15, 2017.}, and flagged them in the dictionary file with the irregular morphological rules we created.

Rules were also added for root words with no morphological change, to assign word features to default (e.g. infinitive) word forms. Additional vocabulary also needed to be added in select cases in which a verb conjugation was the same as a noun. For instance, the word \emph{mercado} (``market'') is not included in the COES dictionary because the verb \emph{mercar} (``to merge'') conjugates to \emph{mercado} in the past participle form. A separate dictionary entry was needed to assign number and gender to the noun sense.

Several additional changes were made to the underlying dictionary and rules derived from COES. First, COES contains a large set of rules for pronominal and transitive enclitic verb endings, since in Spanish it is common to attach clitics to the end of imperative or infinitive verb forms (e.g. \emph{dame} for \emph{da me} or ``give me'', \emph{d{\'a}melo} for \emph{da me lo} or ``give it to me''). The COES dictionary contains the conjugated imperative form with an initial first person singular or singular male enclitic and a flag for converting the enclitic to other pronominal or transitive forms. These word forms are necessary to include in a hash list for spell checking purposes. But when processing text for information extraction, it is easier to first separate all clitics in the target text, then only include rules for verb conjugations but not enclitics, and assign separate person, gender and number features to the detached clitic as a pronoun. The enclitic forms and rules have therefore been removed from the new dictionary and rule files, and a tool for contraction splitting included in the program code.

Second, COES contains the conjugations for several very irregular verbs in the dictionary file, rather than including them in the rules file, if each rule only applies to one verb. This is the case for the verbs \emph{ser}, \emph{estar} (both forms of ``to be''), \emph{ir} (``to go''), \emph{haber} (``to have done''), and \emph{dar} (``to give''). Putting all forms of each of these verbs in the dictionary file takes up less space, if one only needs to know what forms are permitted. But it prevents assigning person, mood, and tense information to each form of those verbs. So for this project, the conjugations of these highly irregular verbs were removed from the dictionary file and individual rules added to the rule file, wherever such cases were found.

Following this dictionary and rule file revision process, a set of tools were written in Python to read in and utilize the lexicon and morphological rules for several purposes arising in event and information extraction processes. Those tools are described in the following sections.

\subsection{Feature Definitions}
\
As in COES, the new dictionary file contains a list of words, one word per line, with each word followed by a ``/'' and one or more letter flags indicating which rules may be applied to create alternative forms of the given root word. The new rules file is a spreadsheet with one morphological rule per line, and the following functional and feature columns:

\begin{enumerate}
\item \textbf{flag:} a letter code to match dictionary words to applicable morphological rules, inherited from COES but with some flags added or removed;
\item \textbf{stem ending:} a regular expression to match to the end of a root word in the dictionary, which matches the portion of the root word to be replaced and, if there is a match, indicates the rule may be applied;
\item \textbf{morph ending:} the characters to insert at the end of the root word in place of the portion that matched the expression in the stem ending column;
\item \textbf{pos:} the part of speech tag for this rule, which may be used as an additional check to make sure the rule should be applied, if other part-of-speech information is available, or may be used to tag words with parts of speech, although only the lexical value and not the syntactic context informs this assignment;
\item \textbf{gender:} male or female, for nouns or adjectives with gender, blank if neutral or not applicable;
\item \textbf{number:} singular or plural, refering to the quantity of countable nouns or adjectives, or to the number of the subject of a verb, depending on the value in the pos column;
\item \textbf{person:} the position of the subject of a verb relative to the speaker (i.e. first, second, or third person), or the person referenced by a pronoun relative to the speaker (also first, second, or third) depending on the value in the pos column;
\item \textbf{mood:} the mood of a verb form (infinitive, indicative, subjunctive, imperative, participle, or gerund);
\item \textbf{tense:} the tense of a verb form (present, past, future, conditional, or imperfect);
\item \textbf{animate:} whether a noun is animate (i.e. person or animal) or inanimate (all others). \emph{Note: COES does not contain animacy information, but one of the morphological rule groups is for turning verbs into participant nouns (e.g. ``command'' into ``commander''), and we added a rule for an ending pattern that reflects persons of certain occupations. Most rules are left blank for this feature, but more animacy-specific rules could be added and matched to animate or inanimate dictionary entries in a later version, especially with the use of additional dictionary resources containing person names.}
\end{enumerate}

\begin{figure*}[h]
\begin{center}
\begin{tabular}{|l|l|l|l|l|l|l|l|l|}
\hline \bf flag & \bf stem end & \bf morph end & \bf pos & \bf gender & \bf number & \bf person & \bf mood & \bf tense \\ \hline
V & ar & o & verb & & singular & first & indicative & present \\
V & (?\textless=\lbrack \^{}cg\rbrack)er & o & verb & & singular & first & indicative & present \\
V & cer & zo & verb & & singular & first & indicative & present \\
V & ger & jo & verb & & singular & first & indicative & present \\
. . . & & & & & & & & \\
 & & & & & & & & \\
S & (?\textless=\lbrack a\rbrack) & s & noun & female & plural &  &  &  \\
S & (?\textless=\lbrack d\rbrack) & es & noun & female & plural &  &  &  \\
S & z & ces & noun & female & plural &  &  &  \\
S & (?\textless=\lbrack eiou'e'o\rbrack) & s & noun & male & plural &  &  &  \\
S & (?\textless=\lbrack 'u'ijlmry\rbrack) & es & noun & male & plural &  &  &  \\
. . . & & & & & & & & \\
 & & & & & & & & \\
 \hline
\end{tabular}
\end{center}
\caption{\label{font-table} Sample of new morphological rule file with suffixes and associated features. }
\end{figure*}

\subsection{Feature Extraction Tool}

As with the COES tools, a user may simply apply the rules in the rule file to the words in the dictionary file to create a list of all possible word forms (and, now, their features). This approach consumes a lot of memory, however, since it would store a feature set for every possible word form. Alternatively, a user could take each target word from a text and compare it directly to entries in the dictionary and rule files, without compiling valid forms first. A user could first search for a root word in the dictionary file that begins with the same characters as the target word, then look up each of the root word's flagged rules, and seeing if applying the corresponding substitution from the rule file produces a word form that matches the target word. This would save memory but take considerably more time for each word to be labeled.

In this project, we've used a hybrid approach which combines the advantages of both of the above options. Instead of loading all possible word forms and their corresponding features up front, the program initially loads only the dictionary and rule files contents. It then labels each target word supplied by the user following the second process described above. It searches the dictionary for the closest root to a target word, using a binary search of a sorted list, then applies morphological rules to successive nearby roots until it finds a match or runs out of roots with the same first letter.

As it selects roots and looks at their morphological forms, the program stores every compiled form in a hashed data structure of word form features for future lookup. For each subsequent labeling request, the function first checks to see if a matching word form is already stored in the word form feature dict, and if not, looks up more roots and their word forms until it finds a match.

Target word forms may have multiple entries with different features, as in the case of \emph{mercado} (``market'' or, less commonly, ``was merged'') described earlier. The functions for feature extraction and lemmatization take an optional part of speech tag, and will only return a feature set that matches the part of speech tag if one is given. The functions also include some default feature preferences based on basic morphological rules, in case multiple matching feature sets are found. The default features for the target word's part of speech and general word ending pattern are returned if no dictionary-based match is found.

\subsection{Lemmatization and Nominalization Tools}

The other two tools created for this project, for use in event and information extraction, are a Spanish-language lemmatizer and a similar tool to convert verbs to a valid nominal form (e.g. \emph{crear} for ``create'' to \emph{creaci\'{o}n} for ``creation'').

The lemmatizer works in a similar fashion to the feature extraction tool described in the last section. It initially searches for a target form's lemma by identifying the closest dictionary roots and trying their morphological rules to find a valid conversion to the target form. Each identified conversion is then stored in a hashed data structure of word forms and their root lemmas for rapid lookup thereafter. The tool checks first to see if each new target word is already stored in the known lemmas dict, so that over time, the lemmatizer becomes faster for looking up common words, but without taking the time or space to load forms that never end up being used.

The nominalization tool is more straight-forward, because it doesn't have to work backwards from a conjugated verb form; it simply converts an infinitive (root) to its rule-specified nominal form. If a target verb isn't yet in infinitive form, the lemmatizer is first applied. The nominalizer then looks up the infinitive in the dictionary file to get the flag for the appropriate nominal conversion rule, and returns the resulting nominal form. For marginal speed savings on common words over time, and for consistency with the other two tools, each verb converted to nominal form is also stored in a nominal form dict for faster lookup if requested again in the future.

\section{Evaluation}

The test data used in the following evaluations is the Spanish language annotated corpus from the CoNLL 2009 Shared Task on Syntactic and Semantic Dependencies in Multiple Languages~\cite{Hajic:09}. The overall objective of the task was to perform and evaluate semantic role labeling (SRL). The CoNLL 2009 Spanish data set was extracted from the AnCora project's AnCora-ES corpus~\cite{Taule:08}, which contains just over 500,000 words of text from Spanish news outlets annotated with syntactic and semantic features and relationships.

Relevant to this project, the data set contains a thorough set of word features for each annotated token, beyond part of speech and lemma, including gender and number for nouns, and person, mood and tense for verbs. These morphological features may be used as part of the input to a supervised machine learning classifier, to obtain the tokens' semantic roles. It follows that one might wish to extract these morphological features independently on new corpora in order to be able to apply a classifier trained on the CoNLL AnCora corpus to label syntactic dependencies and semantic roles in new text.

Since the tools in this project are rule-based, the CoNLL 2009 data set was not used as a formal training set. However, some development-stage tests were necessary to debug the program and fix mislabeled words and features in the dictionary and rule files. This development was performed using the 427,442-word training data file \emph{CoNLL2009-ST-Spanish-train.txt}. To demonstrate thorough lexical and morphological coverage, final evaluation was performed on the 50,368-word development file \emph{CoNLL2009-ST-Spanish-development.txt} and the 1,693-word test file \emph{CoNLL2009-ST-Spanish-trial.txt} combined.

\begin{table}[h]
\begin{center}
\begin{tabular}{|r|c|c|c|}
\hline \bf Feature & \bf Precision & \bf Recall & \bf F-score \\ \hline
Total & 0.930375 & 0.945055 & 0.937658 \\
person & 0.973453 & 0.987022 & 0.980191 \\
mood & 0.974874 & 0.974874 & 0.974874 \\
tense & 0.950213 & 0.968584 & 0.959311 \\
number & 0.955322 & 0.958333 & 0.956826 \\
gender & 0.907974 & 0.910920 & 0.909444 \\
\hline
\end{tabular}
\end{center}
\caption{\label{font-table} Evaluation of word feature extraction performed on CoNLL 2009 Spanish annotated corpus, development and trial files combined. }
\end{table}

Testing was done on all verbs, nouns and adjectives in the annotated corpus. Precision, recall, and f-scores for each word feature are reported in Table 1. All features are above 90\% accurate, and all but gender have precision, recall, and f-scores above 0.95. Gender performs slightly worse at just over 0.9, since it was not distinguished in COES and had to be filled in from scratch for this project. More work could be done to automatically extract genders of irregular nouns from the annotated training set. But this work demonstrates that even without a well-annotated corpus, a basic set of language knowledge and public resources about irregular words can be used to put together a very good word labeler.

We tested the lemmatizer on the same development and trial files from the CoNLL 2009 data set. The CoNLL dataset is also annotated with verbal predicates and their AnCora verb sense, indicating the verb's infinitive. For instance, the token \emph{lleg\'{o}} (``arrived'') is annotated as a predicate of the sense \emph{llegar.b1}, showing its infinitive is \emph{llegar} (``to arrive''). We extracted the verb forms and infinitives of all designated verbal predicates in the data set, ran our lemmatizer on the verb forms, and compared the output to the annotated infinitives. We did not try to detect which verbs would have infinitives annotated in the CoNLL 2009 data set (the task description indicates this is often an arbitrary decision in event extraction, so they provided the predicates for participants). Therefore, only one accuracy score is reported, as opposed to separate precision and recall. Our lemmatizer is again over 90\% accurate for this task.

\begin{table}[h]
\begin{center}
\begin{tabular}{|l|c|c|}
\hline \bf & \bf All verbal & \bf Predicates \\ 
& \bf predicates & \bf not in \\
& & \bf participle form \\ \hline
Total & 5265 & 4929 \\
Correct & 4764 & 4763 \\
Correct/Total & 0.904843 & 0.966322 \\
\hline
\end{tabular}
\end{center}
\caption{\label{font-table} Evaluation of verb lemmatization (conjugated to infinitive form) on CoNLL 2009 Spanish annotated corpus, using verbal predicates annotated in development and trial files combined. }
\end{table}

The most common mistake was past participle forms of verbs, and this does not appear to be due to incorrect lemmatization on our part. The AnCora corpus instead appears to have distinct verb senses for at least some participles. For instance, in several places, the verb form \emph{acusado} (``accused'') is annotated with the sense \emph{acusado.b2} rather than with a sense for the infinitive \emph{acusar} (as our lemmatizer returns), although the word is tagged as a verb and an event predicate rather than an adjectival noun modifier. A simple check for forms ending in \emph{-ado}, \emph{-ido}, or \emph{-echo} (the most common past participle forms) shows that 335 out of 501 words marked incorrect in the test set (or about two thirds) had annotated infinitives in past participle form. Removing word forms with these endings from our lemmatization test increases the accuracy to over 96\%.

\subsection{Discussion}

The materials and tools put together in this project are not the only resources available for morphological analysis of Spanish language text. However, they are straight-forward and easy to use, provide a detailed set of morphological features, leverage the same resources for several useful tasks related to analyzing and linking verb predicates and entity nouns in different forms, and cover a comprehensive general-purpose lexicon much larger than other comparable resources.

This project also demonstrates what might be possible in other languages, if general lexical and morphological resources similar to the COES system are already available for spell checking. The Tron et al paper~\shortcite{Tron:05} introducing \emph{hunmorph} indicates that morphological dictionaries constructed for the Ispell program in other languages are similarly structured. Even if those rule sets do not yet contain information about the features of different morphological forms, it should be easier to assign features to morphological rules that are already written and logically grouped, than to have to assign feature information from scratch to individual items in a lexicon.

Rule-based approaches to word-level analysis can be useful means of leveraging general language knowledge and resources, to maximize the information one has to work with in a language or domain in which higher-level annotated corpora or information extraction tools are less developed. If researchers begin with more detailed general-purpose word-level features, and then apply more inductive and open-ended analysis to learn higher-level semantic structures from their particular corpora of interest, this could also create more comparable foundations for sharing information extraction tools across domains. Processes for inducing event schemas, for instance, that build on general language features should be easier to share and replicate than supervised approaches, even if specific events and entities will  ultimately vary with the context from which they are derived.

\bibliography{eacl2017}
\bibliographystyle{eacl2017}

\end{document}